\documentclass{article}

\usepackage[preprint]{neurips_2025}

\usepackage[utf8]{inputenc} % allow utf-8 input
\usepackage[T1]{fontenc}    % use 8-bit T1 fonts
\usepackage{hyperref}       % hyperlinks
\usepackage{url}            % simple URL typesetting
\usepackage{booktabs}       % professional-quality tables
\usepackage{amsfonts}       % blackboard math symbols
\usepackage{amsmath}        % for advanced math symbols
\usepackage{amssymb}        % for more math symbols
\usepackage{nicefrac}       % compact symbols for 1/2, etc.
\usepackage{microtype}      % microtypography
\usepackage{xcolor}         % colors
\usepackage{appendix}
\usepackage{graphicx}
\usepackage{natbib}         % Ensure natbib is loaded
\usepackage{tikz}
\usetikzlibrary{shapes.geometric, arrows.meta, positioning}

\title{Through a Steerable Lens: Magnifying Neural Network Interpretability via Phase-Based Extrapolation}

\author{%
  Farzaneh Mahdisoltani\textsuperscript{\,1,2} \quad
  Saeed Mahdisoltani \textsuperscript{\,3} \quad
  Roger B. Grosse\textsuperscript{\,1,2} \quad 
  David J. Fleet\textsuperscript{\,1,2} \quad \\
  \textsuperscript{1}University of Toronto \quad
  \textsuperscript{2}Vector Institute \quad
  \textsuperscript{3} MIT\\
}
\begin{document}

\maketitle

\begin{abstract}
Understanding the internal representations and decision mechanisms of deep neural networks remains a critical open challenge. While existing interpretability methods often identify influential input regions, they may not elucidate \textit{how} a model distinguishes between classes or \textit{what specific changes} would transition an input from one category to another. To address these limitations, we propose a novel framework that visualizes the implicit “path” between classes by treating the network gradient as a form of infinitesimal motion. Drawing inspiration from phase-based motion magnification, we first decompose images using invertible transforms—specifically the Complex Steerable Pyramid—then compute class-conditional gradients in the transformed space. Rather than iteratively integrating the gradient to trace a full path, we amplify the one-step gradient to the input and perform a linear extrapolation to expose how the model “moves” from source to target class. By operating in the steerable-pyramid domain, these amplified gradients produce semantically meaningful, spatially coherent morphs that highlight the classifier’s most sensitive directions, giving insight into the geometry of its decision boundaries. Experiments on both synthetic and real-world datasets demonstrate that our phase-focused extrapolation yields perceptually aligned, semantically meaningful transformations, offering a novel, interpretable lens into neural classifiers’ internal representations.
\end{abstract}

\section{Introduction}

Deep neural networks (DNNs) have achieved remarkable success across numerous domains, especially in computer vision~\citep{krizhevsky2012imagenet, he2016deep}. However, their complex, hierarchical architectures and vast parameter spaces render them difficult to interpret, often leading to their characterization as "black boxes." This lack of transparency raises concerns around trust, debugging, fairness, and robustness, prompting significant interest in developing methods for model interpretability~\citep{doshi2017towards, rudin2019stop}.

A core challenge in interpretability lies in understanding how a model internally represents and distinguishes between classes. Many existing methods attempt to attribute model predictions to specific input features. Saliency maps~\citep{simonyan2013deep, springenberg2014striving, smilkov2017smoothgrad, sundararajan2017axiomatic, selvaraju2017grad} highlight influential input regions but often lack spatial coherence and do not describe how changes in those features affect classification outcomes. Adversarial perturbations~\citep{szegedy2014intriguing, goodfellow2014explaining} explore model sensitivity by finding minimal input modifications that alter predictions, yet such perturbations are typically imperceptible and semantically uninformative. Counterfactual explanations~\citep{wachter2017counterfactual, goyal2019counterfactual} aim to identify minimal meaningful changes that flip model predictions, but generating realistic and interpretable counterfactuals remains challenging.

In this work, we propose a complementary lens on interpretability: 
Rather than highlighting input importance, we visualize the trajectory that a model implicitly perceives between one class and another. 
This idea is inspired by phase-based motion magnification~\citep{wu2012eulerian}, where imperceptibly small temporal displacements in a video are amplified to produce visible motion without introducing artificial content. 
We draw a conceptual parallel to motion magnification: the gradient of a neural network with respect to an input image is analogous to subtle motion—an infinitesimal perturbation that nudges the image toward a particular target class. Due to its infinitesimal nature, this gradient is typically not visually discernible. However, by analogy to subtle motions in video that can be revealed through temporal magnification techniques, we hypothesize that it is possible to amplify these gradients in a way that makes them perceptually meaningful.
Hence, we treat the classifier’s gradient as a latent, structurally meaningful signal that becomes interpretable when magnified along semantically aligned axes in a structured domain. To realize this, we move beyond raw pixel space and operate within a transformed domain that encodes spatially localized, multi-scale, and orientation-selective features. The complex steerable pyramid~\citep{freeman1991design} provides such a decomposition, separating local image structure into amplitude (feature strength) and phase (feature position). This representation enables manipulation in perceptually aligned subspaces and is more disentangled than the pixel basis. 

Importantly, we do not iteratively integrate the gradient to trace a full path between classes. Rather, we amplify a single local tangent vector—the immediate direction of change implied by the gradient—creating a linear extrapolation that reveals how the model perceives the transition from source to target class in its feature space.
This formulation offers new insights into model decision boundaries. Because the steerable pyramid represents localized, disentangled image structures, gradients in this space yield semantically meaningful and spatially coherent transformations. Our method highlights the directions in feature space along which the classifier is most sensitive, offering a dynamic view of model behavior grounded in perceptual structure.

\paragraph{Contributions.} 
We summarize our main contributions as follows:
\begin{itemize}
    \item We introduce a novel framework for visualizing neural network sensitivities by extrapolating image gradients in transformed phase-amplitude spaces (Fourier and Steerable Pyramid).
    \item We show that extrapolating phase components yields perceptually coherent and semantically meaningful image morphs between classes.
    \item We derive principled update rules using Wirtinger calculus for manipulating complex coefficients in the transformed domain.
    \item We demonstrate that our approach reveals structured decision trajectories and disentangled feature directions in both synthetic and real-world datasets.
\end{itemize}

\section{Related Work}
\label{sec:related_work}

Our work connects to several lines of research in machine learning interpretability and signal processing.

\paragraph{Saliency Methods and Attribution.} This is perhaps the most common approach to interpretability. Methods like Vanilla Gradients \citep{simonyan2013deep}, Guided Backpropagation \citep{springenberg2014striving}, SmoothGrad \citep{smilkov2017smoothgrad}, Integrated Gradients \citep{sundararajan2017axiomatic}, and Grad-CAM \citep{selvaraju2017grad} aim to assign importance scores to input features (usually pixels) based on their contribution to the output. While valuable for identifying relevant regions, these methods produce static heatmaps that often suffer from noise \citep{smilkov2017smoothgrad} and primarily indicate \textit{where} the model looks, not \textit{how} feature changes affect the decision or the structure of the decision boundary itself. Our method differs by generating dynamic sequences illustrating \textit{transformations} between classes.

\paragraph{Adversarial Examples and Counterfactual Explanations.} Adversarial examples \citep{szegedy2014intriguing, goodfellow2014explaining} find small input perturbations that cause misclassification. They reveal model vulnerabilities and local properties of the decision boundary but typically lack semantic interpretability. Counterfactual explanations \citep{wachter2017counterfactual} seek minimal, meaningful input changes that alter the outcome. Methods like \citep{goyal2019counterfactual} aim for visual counterfactuals, often requiring generative models or complex optimization objectives to ensure realism and interpretability. Our approach also generates input modifications leading to a different class, but does so by following a gradient-derived path in a structured feature space, aiming for inherent perceptual coherence without explicit generative modeling during the morphing process.

\paragraph{Multiscale Representations and Signal Processing.} Our use of Fourier and steerable pyramid transforms draws heavily from classical signal processing. The Fourier transform represents signals via global sinusoids, where phase encodes position \citep{oppenheim1999discrete}. Steerable pyramids \citep{freeman1991design, simoncelli1995steerable} provide a localized, multi-scale, multi-orientation decomposition, akin to wavelet transforms \citep{mallat1999wavelet}, but with better orientation selectivity and often implemented with complex coefficients that separate local amplitude and phase \citep{portilla2000parametric}. The insight that manipulating phase in such domains can reveal or amplify subtle structural changes or motion is central to Eulerian Video Magnification \citep{wu2012eulerian, wadhwa2013phase, wadhwa2016phase}. We adapt this idea, using \textit{classifier gradients} to drive phase manipulation for the purpose of interpretability, rather than amplifying existing temporal variations.

\paragraph{Gradient Extrapolation and Feature Space Interpretation.} Some works explore model behavior by perturbing inputs along gradient directions \citep{simonyan2013deep} or optimizing inputs to maximize class scores (activation maximization). Others attempt to interpret or manipulate representations in intermediate layers of the network \citep{olah2017feature, bau2017network}. Recent work has also considered extrapolating gradients in learned feature spaces to understand sensitivities \citep{ghorbani2019interpretation}. However, gradients in pixel space can be noisy, and gradients in learned feature spaces can be hard to interpret if the features themselves are entangled or lack clear semantics. Our approach differs by performing gradient extrapolation in a \textit{fixed, interpretable, structured transform space} (FT or CSP), leveraging the properties of amplitude and phase in these domains to achieve interpretable visualizations.

\paragraph{Summary.} Our method uniquely combines gradient-based sensitivity analysis with insights from multiscale signal processing. By extrapolating class-conditional gradients primarily in the phase domain of Fourier or steerable pyramid representations, we generate dynamic, coherent visualizations of inter-class transitions, offering a distinct perspective compared to static saliency or standard adversarial/counterfactual methods.

\section{Method}
\label{sec:method}

We propose generating interpretable trajectories between classes by extrapolating gradients in a transformed amplitude-phase space. Let $s(\mathbf{x})$ be an input image defined over spatial coordinates $\mathbf{x}$. Our framework involves transforming $s$, computing gradients in the transform domain, extrapolating, and reconstructing.

\subsection{General Framework}

We begin by applying an invertible transformation \(\mathcal{F}\) to the input image \(s(\mathbf{x})\) to obtain its representation in a different domain,  $S(\boldsymbol{\omega}) = \mathcal{F}(s(\mathbf{x}))$,
where \(\boldsymbol{\omega}\) denotes the coordinates in the transformed domain (e.g., frequency or spatial position, scale, orientation). 
To discover the direction in transform space that maximally increases the likelihood of a target class \(t^*\), we define a loss and compute its gradient with respect to the transformed domain coefficients. 
% \begin{align}
%     &\mathcal{L}\bigl(\mathcal{N}(\mathcal{F}^{-1}(S(\boldsymbol{\omega}))),\,t^*\bigr)\\
%     &\mathbf{g} \;=\;\nabla_{S(\boldsymbol{\omega})}\,\mathcal{L} \;=\;\frac{\partial \mathcal{L}}{\partial S(\boldsymbol{\omega})},
% \end{align}
\begin{equation}
    \mathcal{L}\bigl(\mathcal{N}(\mathcal{F}^{-1}(S(\boldsymbol{\omega}))),\,t^*\bigr) \qquad \text{and} \qquad
    \mathbf{g} =\nabla_{S(\boldsymbol{\omega})}\,\mathcal{L} =\frac{\partial \mathcal{L}}{\partial S(\boldsymbol{\omega})},
\end{equation}

\begin{figure}
    \centering
    {\includegraphics[width=0.95\textwidth]{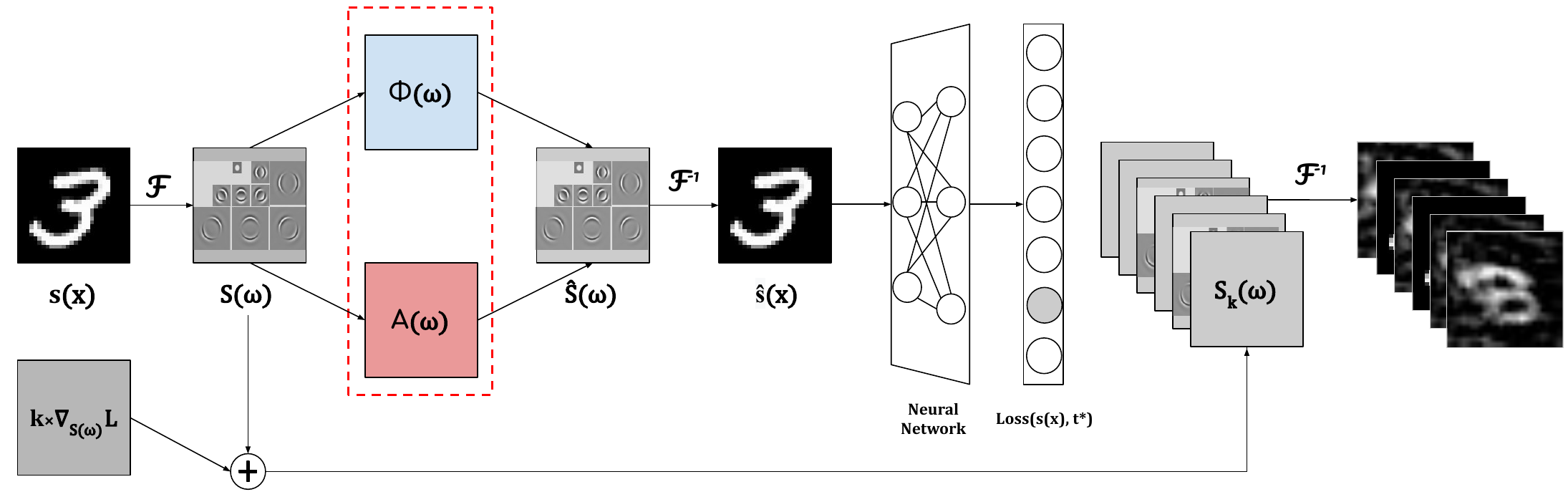}}\\
    \caption{Block diagram of the phase-based gradient extrapolation pipeline. An input image \(s(\mathbf{x})\) is mapped through \(\mathcal{F}\) to obtain transformed parameterization \(S(\boldsymbol{\omega})\), which are split into phase \(\Phi(\boldsymbol{\omega})\) and amplitude \(A(\boldsymbol{\omega})\). Gradients in this domain linearly extrapolate \(S(\boldsymbol{\omega})\) to generate a series of modified coefficients \(S_{k}(\boldsymbol{\omega})\). The inverse transform \(\mathcal{F}^{-1}\) produces the morphed images \(\hat{s}_{k}(\mathbf{x})\).}
    \label{fig:evm_diagram}
\end{figure}

 We then extrapolate from the initial representation \(S_0(\boldsymbol{\omega}) = \mathcal{F}(s(\mathbf{x}))\) along this gradient to generate a sequence \(\{S_k(\boldsymbol{\omega})\}\) with refinements that respect the amplitude–phase decomposition; see Sec.~\ref{subsec:gradient_amp_phase}).
\begin{equation}
    S_k(\boldsymbol{\omega}) = S_0(\boldsymbol{\omega}) + k\,\alpha\,\mathbf{g},
\end{equation}
 Finally, each extrapolated coefficient set \(S_k(\boldsymbol{\omega})\) is mapped back to the image domain as  
%\begin{equation}
    $s_k(\mathbf{x}) = \mathcal{F}^{-1}(S_k(\boldsymbol{\omega}))$,
%\end{equation}
yielding a sequence of images \(\{s_k(\mathbf{x})\}_{k=0,1,\dots}\) that visualizes the morphing trajectory toward the target class \(t^*\). This forms the foundation for our method of extrapolating model gradients in transformed space to visualize semantic trajectories (see Figure~\ref{fig:evm_diagram} for the overall framework).

\subsection{Choice of Transformation $\mathcal{F}$.}
The choice of \(\mathcal{F}\) significantly influences the interpretability and granularity of classifier‐morph visualizations. A transform domain must explicitly encode local structural properties such as orientation, scale, and spatial positioning to support meaningful manipulations. In the Fourier domain, phase carries the bulk of perceptual information and encodes local positional information with high sensitivity \citep{KuglinHines1975, oppenheim1981importance, oppenheim1999discrete}.
The phase component is therefore particularly effective for detecting and manipulating fine‐scale structural shifts and motion information \citep{fleet1990, BarronFleet1992}. Below we contrast two complementary decompositions:
\begin{itemize}
    \item \textbf{Discrete Fourier Transform (DFT):} $S(\boldsymbol{\omega})$ represents global frequency components (magnitude and phase) \citep{oppenheim1999discrete}. Phase shifts in the DFT domain primarily correspond to global translations of the image content \citep{BarronFleet1992}. Extrapolating DFT phase reveals the network's sensitivity to global spatial arrangement.
    \item \textbf{Complex Steerable Pyramid (CSP):} $S(\boldsymbol{\omega})$ represents the image in terms of localized basis functions (wavelets) \citep{FleetJepsonJenkin1991}, selective for specific spatial locations, scales, and orientations. $A(\boldsymbol{\omega})$ represents the local energy-contrast of that feature, while $\Phi(\boldsymbol{\omega})$ represents its local phase-position \citep{simoncelli1995steerable}. Phase shifts in the CSP domain correspond to local shifts or deformations of oriented structures. This localized, oriented nature aligns well with features learned by CNNs \citep{krizhevsky2012imagenet}. Extrapolating CSP phase can thus reveal how the network relies on fine-grained, localized structural changes to differentiate classes.
\end{itemize}

While the DFT offers a compact global view, its lack of localization limits interpretability.  The CSP’s tight‐frame, orientation‐selective decomposition, by contrast, aligns more directly with both human perception and CNN feature geometry—so we generally favor it for our visualizations.

\subsection{Gradient Extrapolation in Amplitude-Phase Space}
\label{subsec:gradient_amp_phase}
Our choice of transformations induces a complex-valued parameterization of the input image in the transformed domain, which we denote by $\boldsymbol{\omega}$. Accordingly, we employ Wirtinger calculus to perform principled differentiation with respect to complex-valued variables, enabling the computation of class-conditional gradients in the transformed domain.
% , and extrapolate primarily along the phase direction.
% In order to compute the gradient of a real-valued loss $\mathcal{L}$ with respect to the complex coefficients $S(\boldsymbol{\omega})$. 
Since $\mathcal{L}$ is a real-valued function and $S(\boldsymbol{\omega})$ is complex, the direction of steepest ascent is given by the derivative with respect to the complex conjugate, $S^*(\boldsymbol{\omega})$, using Wirtinger calculus \citep{brandwood1983complex}. Let $\mathbf{g}^*$ denote this gradient vector:
\begin{equation}
    \mathbf{g}^* = \nabla_{S^*(\boldsymbol{\omega})} \mathcal{L} = \frac{\partial \mathcal{L}}{\partial S^*(\boldsymbol{\omega})}.
\end{equation}
(See Appendix \ref{appendix:wirtinger_derivation} for details on Wirtinger calculus.)

Inspired by phase-based motion magnification~\citep{wu2012eulerian, wadhwa2013phase}, we depart from the direct application of the complex conjugate gradient $\mathbf{g}^*$ to the transformed signal $S(\boldsymbol{\omega})$. Instead, we analyze the impact of this gradient in a more interpretable parameterization by decomposing each coefficient into its local amplitude and phase components. Specifically, each complex coefficient is expressed as
\begin{equation}\label{eq:amp_phase_decomp}
S(\boldsymbol{\omega}) = A(\boldsymbol{\omega})\,e^{j\Phi(\boldsymbol{\omega})},
\end{equation}
where \(A(\boldsymbol{\omega}) = |S(\boldsymbol{\omega})|\) represents the amplitude and \(\Phi(\boldsymbol{\omega}) = \arg(S(\boldsymbol{\omega}))\) captures the 
% local structural 
phase.

The amplitude–phase decomposition underlies our method: amplitude captures local feature strength while phase encodes structure, position, or alignment~\citep{fleet1990, oppenheim1981importance, portilla2000parametric}, and prior work has shown that subtle phase modifications can induce significant, spatially coherent changes in the reconstructed signal~\citep{wu2012eulerian, wadhwa2016phase}. 
Using the chain rule within the Wirtinger framework (detailed derivation in Appendix \ref{appendix:extrapolation_derivation}), the gradient $\mathbf{g}^* = \nabla_{S^*(\boldsymbol{\omega})} \mathcal{L}$ can be related to the gradients with respect to amplitude and phase:
\begin{equation} \label{eq:wirtinger_chain_rule}
\nabla_{S^*(\boldsymbol{\omega})} \mathcal{L} = \frac{1}{2} \left( \nabla_{A(\boldsymbol{\omega})} \mathcal{L} \frac{S(\boldsymbol{\omega})}{A(\boldsymbol{\omega})} + j \nabla_{\Phi(\boldsymbol{\omega})} \mathcal{L} \frac{S(\boldsymbol{\omega})}{A(\boldsymbol{\omega})^2} \right),
\end{equation}
% where $\nabla_{A(\boldsymbol{\omega})} \mathcal{L} = \frac{\partial \mathcal{L}}{\partial A(\boldsymbol{\omega})}$ and $\nabla_{\Phi(\boldsymbol{\omega})} \mathcal{L} = \frac{\partial \mathcal{L}}{\partial \Phi(\boldsymbol{\omega})}$.
% 
Performing a single gradient ascent step (with step size $\alpha$) we compute $S_{k+1} = S_k + \alpha \nabla_{S_k^*} \mathcal{L}$, yielding
\begin{align} \label{eq:multiplicative_update}
S_{k+1}(\boldsymbol{\omega}) 
&= S_k(\boldsymbol{\omega}) \left( 1 + \frac{\alpha}{2} \frac{\nabla_{A_k(\boldsymbol{\omega})} \mathcal{L}}{A_k(\boldsymbol{\omega})} + \frac{j\alpha}{2} \frac{\nabla_{\Phi_k(\boldsymbol{\omega})} \mathcal{L}}{A_k(\boldsymbol{\omega})^2} \right). 
\end{align}

% \begin{align} \label{eq:complex_update}
% S_{k+1}(\boldsymbol{\omega}) &= S_k(\boldsymbol{\omega}) + \alpha \nabla_{S_k^*(\boldsymbol{\omega})} \mathcal{L} \\
% &= S_k(\boldsymbol{\omega}) \left( 1 + \frac{\alpha}{2} \frac{\nabla_{A_k(\boldsymbol{\omega})} \mathcal{L}}{A_k(\boldsymbol{\omega})} + \frac{j\alpha}{2} \frac{\nabla_{\Phi_k(\boldsymbol{\omega})} \mathcal{L}}{A_k(\boldsymbol{\omega})^2} \right). \label{eq:multiplicative_update}
% \end{align}
Let the complex term in the parenthesis be $\mathbf{z}_k = |\mathbf{z}_k| e^{j\arg(\mathbf{z}_k)}$. Then the update can be written in terms of amplitude and phase as:
\begin{align}
\Phi_{k+1}(\boldsymbol{\omega}) &= \Phi_k(\boldsymbol{\omega}) + \arg(\mathbf{z}_k)\\ 
A_{k+1}(\boldsymbol{\omega}) &= |\mathbf{z}_k| A_k(\boldsymbol{\omega}).
\end{align}
This shows that a gradient step simultaneously scales the amplitude and shifts the phase. The phase shift induced by a single step (calculated at $k=0$ using $S_0, A_0, \Phi_0$) is:
\begin{equation} \label{eq:delta_phi}
\Delta \Phi(\boldsymbol{\omega}) = \arg\left( 1 + \frac{\alpha}{2} \frac{\nabla_{A_0(\boldsymbol{\omega})} \mathcal{L}}{A_0(\boldsymbol{\omega})} + \frac{j\alpha}{2} \frac{\nabla_{\Phi_0(\boldsymbol{\omega})} \mathcal{L}}{A_0(\boldsymbol{\omega})^2} \right).
\end{equation}

\paragraph{Phase-Focused Extrapolation.}

 Representations such as the Fourier transform and the complex steerable pyramid~\citep{freeman1991design} decompose images into basis functions aligned with specific scales and orientations. However, it is the phase component of these decompositions that plays a critical role in capturing fine-grained positional information. While amplitude coefficients encode the energy or strength of features, phase encodes their spatial alignment—shifting phase alters the local position of image structures without changing their frequency or orientation content~\citep{oppenheim1999discrete}.

Inspired by the significant perceptual impact of phase manipulations \citep{wadhwa2016phase} and empirical observations that phase changes drive the most coherent structural transformations in our setting, we adopt a simplified extrapolation strategy. We compute the initial phase gradient direction $\Delta \Phi(\boldsymbol{\omega})$ using Eq.~\eqref{eq:delta_phi} (typically with an amplifier $\alpha$ determining the overall magnitude of change per step) and perform linear extrapolation purely in the phase domain, keeping the amplitude constant:
\begin{equation}
\Phi_k(\boldsymbol{\omega}) = \Phi_0(\boldsymbol{\omega}) + k \cdot \Delta \Phi(\boldsymbol{\omega}) \qquad \text{and} \qquad
A_k(\boldsymbol{\omega}) = A_0(\boldsymbol{\omega}). \label{eq:amp_const_rule}
\end{equation}\label{eq:phase_extrap_rule}
The extrapolated coefficients are then reconstructed:
\begin{equation} \label{eq:reconstruct_phase_extrap}
S_k(\boldsymbol{\omega}) = A_0(\boldsymbol{\omega}) e^{j\Phi_k(\boldsymbol{\omega})} \qquad\text{and}\qquad  s_k(\mathbf{x}) = \mathcal{F}^{-1}(S_k(\boldsymbol{\omega}))
\end{equation}
This phase-focused approach isolates the structural changes guided by the gradient, leading to clearer visualizations of how the network distinguishes classes based on form and position rather than just contrast.
\section{Experimental Results}
\label{sec:experiments}
We evaluate our method on a synthetic dataset designed for controlled analysis and on the standard MNIST \citep{lecun1998gradient} benchmark, as well as FER2013 dataset of real-world facial expressions \cite{goodfellow2013fer2013}.
% demonstrating the ability of phase-based extrapolation to generate interpretable class transformations. 
We primarily use the CSP with 4 orientations and 3 scales as the transform, unless otherwise noted. Gradients are computed using PyTorch autograd, which implicitly handles complex variables via Wirtinger calculus principles, by treating real and imaginary parts separately for backpropagation. All experiments can be reproduced on a standard local machine or in a free Google Colab notebook without any code changes.
%through the transform  $\mathcal{F}$ and the network.
\subsection{Arcade Synthetic Dataset}
\begin{figure}[htbp]
    \centering
    {\includegraphics[width=1\textwidth]{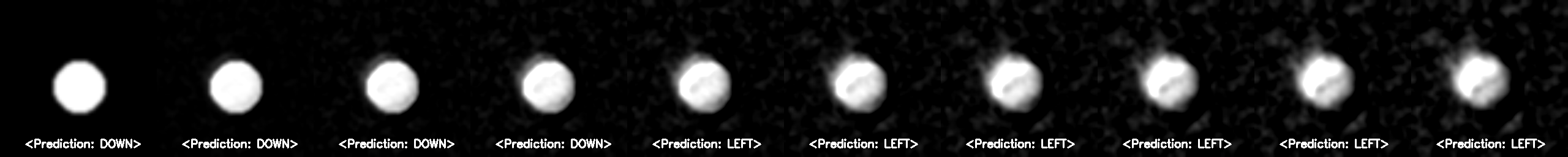}}\\
    {\includegraphics[width=1\textwidth]{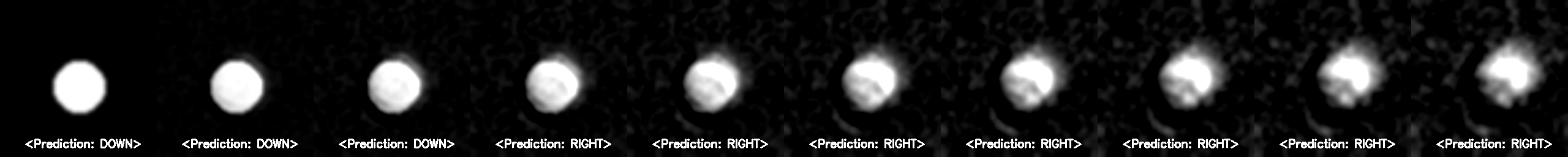}} \\
    {\includegraphics[width=1\textwidth]{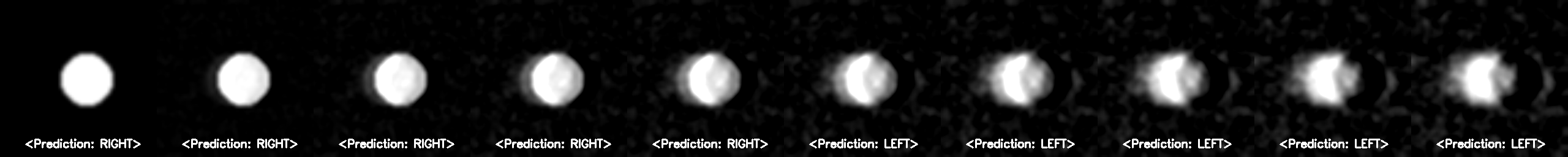}} \\
    {\includegraphics[width=1\textwidth]{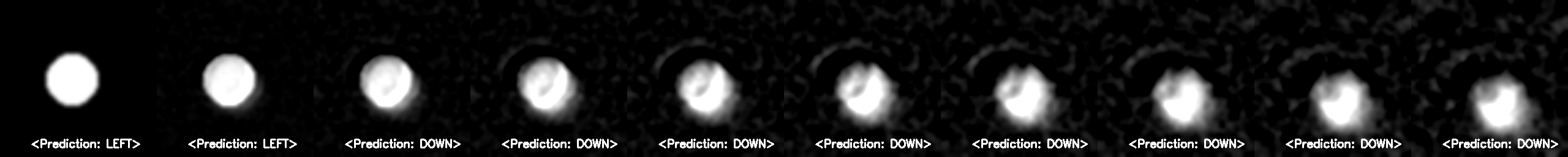}} \\
    \caption{Morphing sequences of the Arcade dataset using CSP extrapolation. Each row visualizes the trajectory from source (leftmost frame in each row) to target direction. The morphing shows the object moves smoothly in the intended direction. From top to bottom, the rows transform: (Down → Left), (Down → Right), (Right → Left), and (Left → Down).}
    \label{fig:arcade_results}
\end{figure}

The Arcade dataset consists of grayscale images ($64 \times 64$) showing a white disk against a black background. The disk is displaced from the center in one of four cardinal directions (Up, Down, Left, Right) by a small, random amount. The task is to classify the direction of displacement. This simple setup allows us to precisely understand the expected transformations between classes.
We trained a simple CNN classifier with two convolutional layers followed by two fully connected layers. We apply our method to morph images between different directional classes (See Figure \ref{fig:arcade_results}). The generated trajectories exhibit meaningful motion of the disk in the target direction, reflecting the underlying change required to switch class labels according to the trained model. This confirms that phase extrapolation along the gradient direction in CSP space successfully captures and visualizes the network's sensitivity to the object's spatial position, the defining characteristic of this dataset. The motion appears natural and directly corresponds to the semantic difference between the classes.
The disk’s distortion is a feature, not an artifact:
Our visualization shows that the classifier values the motion of the center of mass, not shape fidelity. The gradient points to where mass should move to raise the target class score; it never insists the white region stay circular. Any diffused shape that keeps the same centroid—and can shear, stretch, or blur along the way—satisfies the gradient.
By revealing this preference, the visualization achieves its goal of showing what the network cares about (centroid), and what it ignores (precise contour).

% \textbf{Object Deformation.} It is noticeable that the object deforms, while moving towards the target direction. This is rooted in the characteristics of our visualization method, which amplifies the classifier gradient, not a geometric constraint. The gradient indicates how the class score changes when mass is displaced; it does not encode that the white region must remain circular. In other words, any shape whose center-of-mass follows the same path will satisfy the gradient, so the visualized object is free to shear, elongate, or otherwise deform while moving. What matters is that the motion aligns with the direction of steepest ascent in class confidence, not that the shape preserves Euclidean symmetry. Consequently, the circle in \ref{fig:arcade_results} drifts and distorts: a perfectly acceptable outcome that reveals the model’s true sensitivity, rather than an artifact to “correct.”

\subsection{MNIST Dataset}
\begin{figure}[htbp] 

    \centering
    {\includegraphics[width=1\textwidth]{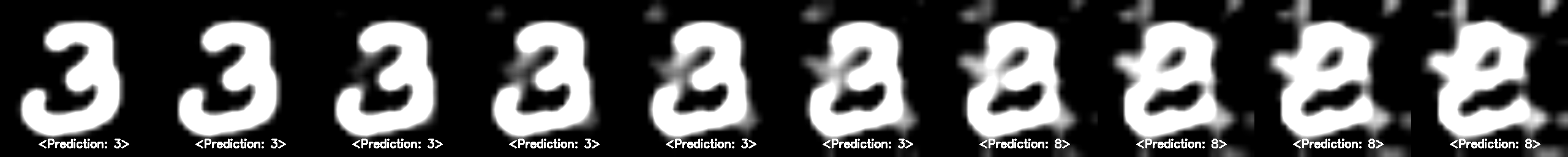}}\\
    {\includegraphics[width=1\textwidth]{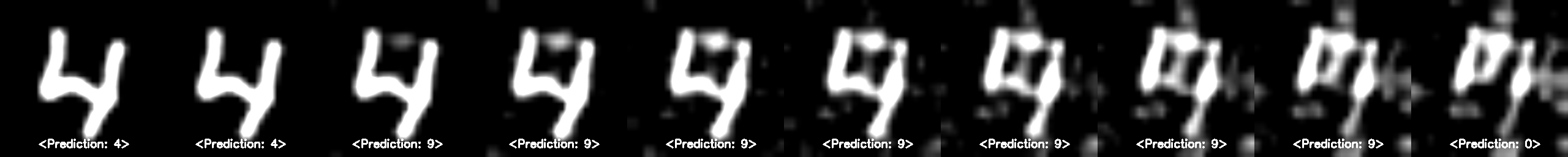}}\\
    {\includegraphics[width=1\textwidth]{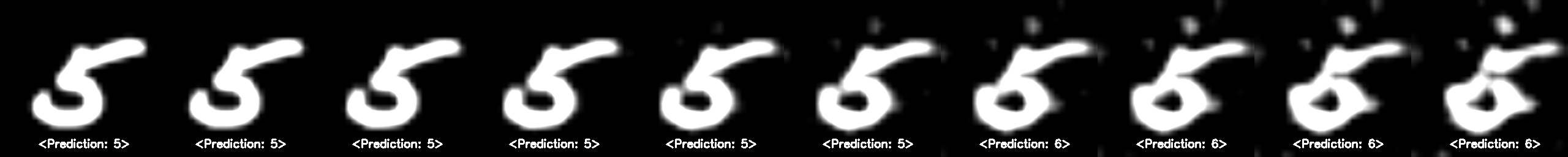}}\\
    {\includegraphics[width=1\textwidth]{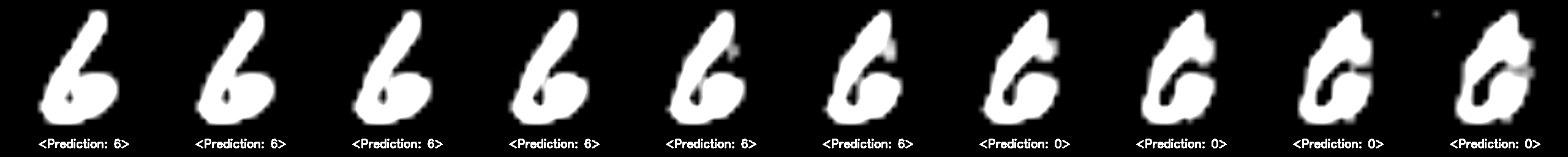}}\\
    {\includegraphics[width=1\textwidth]{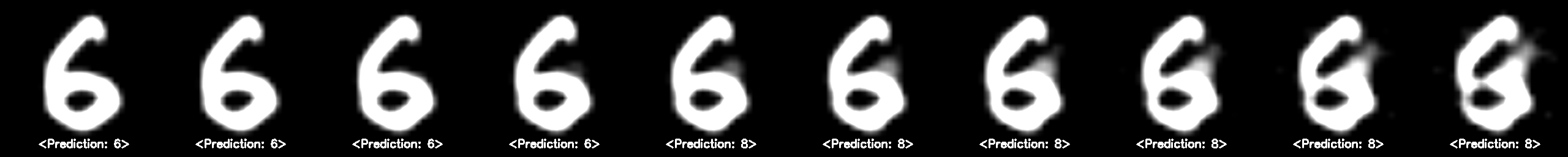}}\\
    {\includegraphics[width=1\textwidth]{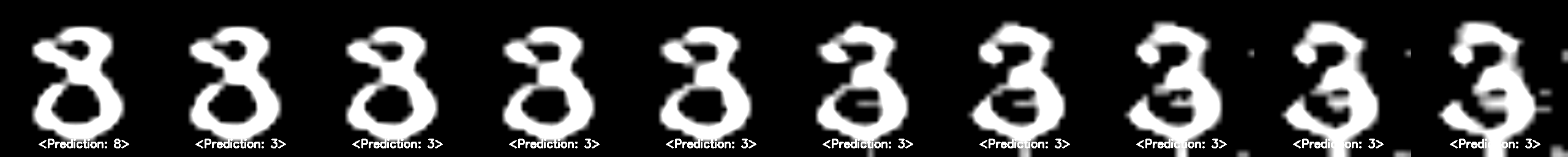}}\\
    {\includegraphics[width=1\textwidth]{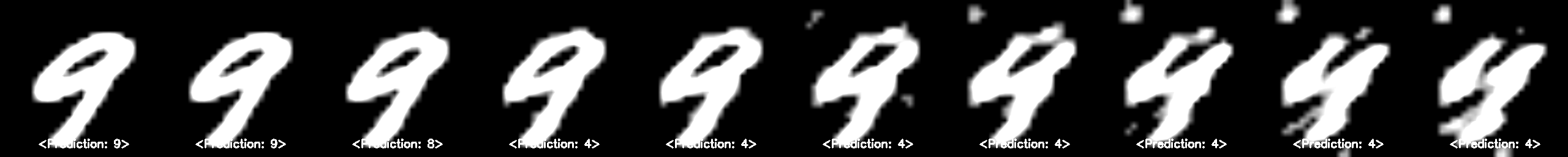}}\\
    \caption{Transformation trajectories on MNIST: Each row shows a source digit (left frame) morphing towards a target digit by following the gradients in CSP domain. The intermediate frames highlight class-relevant features like stroke curvature, closures, and intersections, as learned by the classifier. From top to bottom, the rows transform: (3 → 8), (4 → 9), (5 → 6), (6 → 8), (8 → 3),and (9 → 4)}
    \label{fig:mnist_results1}
\end{figure}

\begin{figure}[htbp]
    \centering
    \includegraphics[width=1\textwidth, trim=0 0 0 20, clip]{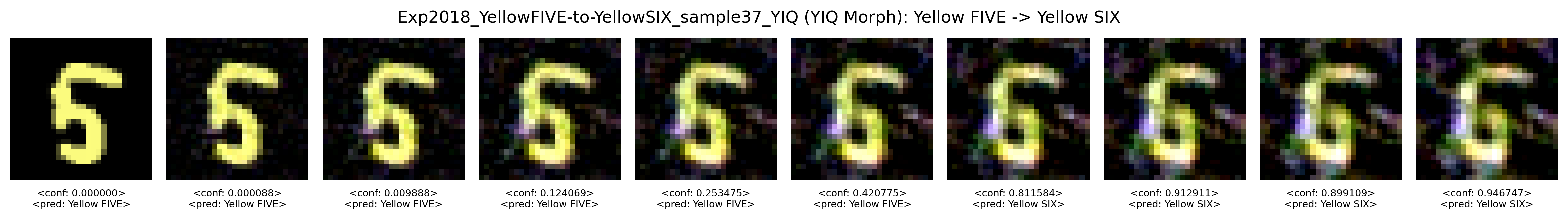}\vspace{-4mm}
    \includegraphics[width=1\textwidth, trim=0 0 0 20, clip]{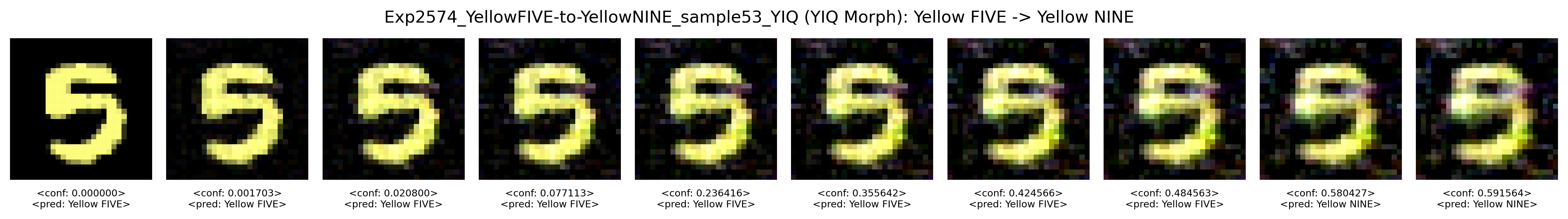}\vspace{-4mm}
    \includegraphics[width=1\textwidth, trim=0 0 0 20, clip]{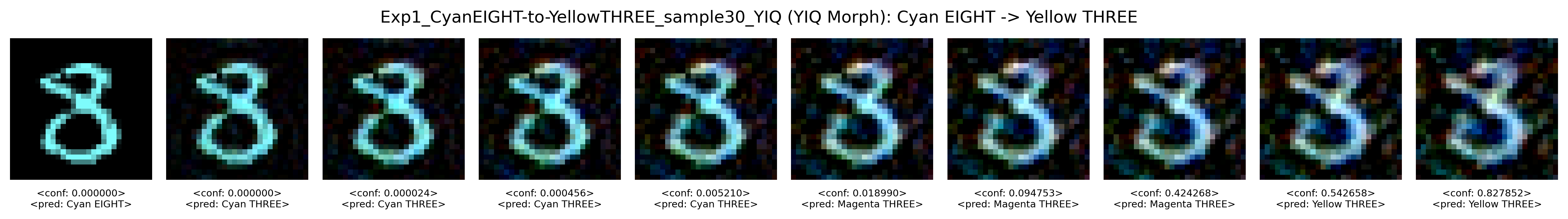}\vspace{-4mm}
    \includegraphics[width=1\textwidth, trim=0 25 0 20, clip]{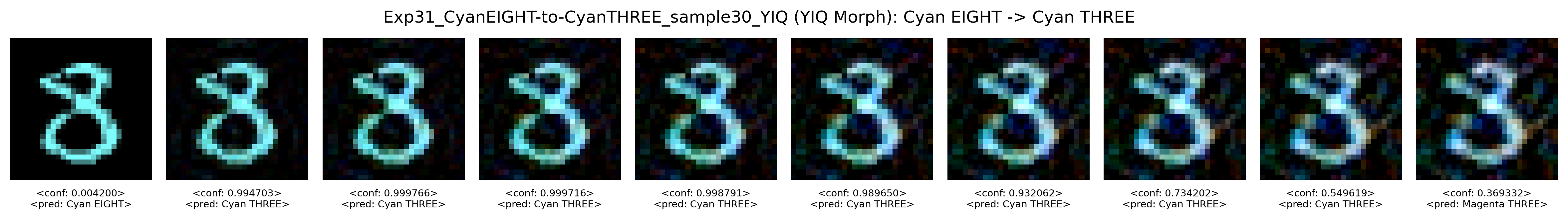}\vspace{-1mm}
   \caption{Transformation trajectories on MNIST with \emph{composite} colors in the YIQ representation.  
   Each row depicts a digit initially colored in a two‐channel hue (magenta, yellow, or cyan) morphing to its target via phase extrapolation in each channel independently. Updating Y drives shape changes while I and Q updates steer color shifts, demonstrating joint shape-color control. From top to bottom, the rows transform: (5 → 6), (5 → 9), (5 → 8), and (8 → 3).}
   \label{fig:color_mnist_2}
\end{figure}

We used the standard MNIST dataset \citep{lecun1998gradient} of handwritten digits ($28 \times 28$). We trained a LeNet-like architecture \citep{lecun1998gradient} consisting of two convolutional layers followed by two fully connected layers to achieve high classification accuracy.
We selected source images correctly classified by the model and generated morphing sequences towards different target digit classes using our method. Figures \ref{fig:mnist_results1}, \ref{fig:color_mnist_1}, \ref{fig:color_mnist_2} show several examples. The results are compelling:
\begin{itemize}
    \item The transformations are visually smooth and temporally coherent.
    \item The changes are semantically meaningful and often align with human intuition about how digits relate. For instance, a '3' morphing into an '8' gradually closes its top and bottom openings. A '5' transforming into a '6' develops a closed loop at the bottom. 
    \item These visualizations directly illustrate the specific structural features the network modifies to transition between class representations. The localized nature of the CSP allows the changes to be spatially focused (e.g., modifying a specific stroke or loop).
\end{itemize}
Compared to naive pixel-space gradient ascent (which often adds noise or globally brightens/darkens) or Fourier-based extrapolation (which tends to produce more global shifts or blurring, see Appendix \ref{appendix:fourier_results}), our method yields significantly more interpretable and structurally relevant transformations.

\subsection{Extension to Color MNIST}
\begin{figure}[htbp]
    \centering
    \includegraphics[width=1\textwidth, trim=0 0 0 20, clip]{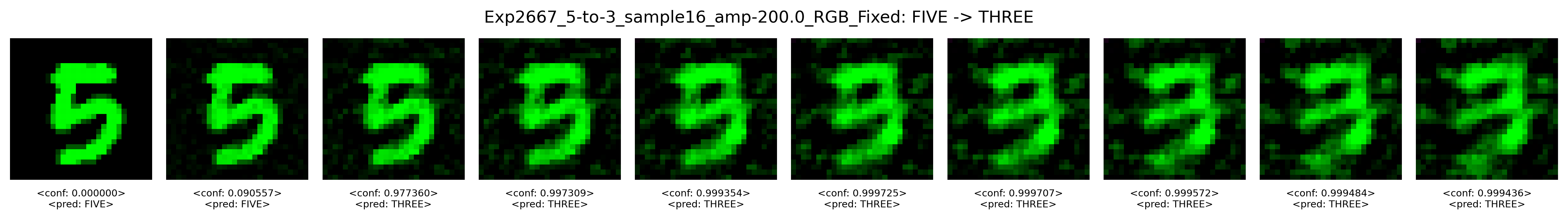}\vspace{-4mm}
    \includegraphics[width=1\textwidth, trim=0 0 0 20, clip]{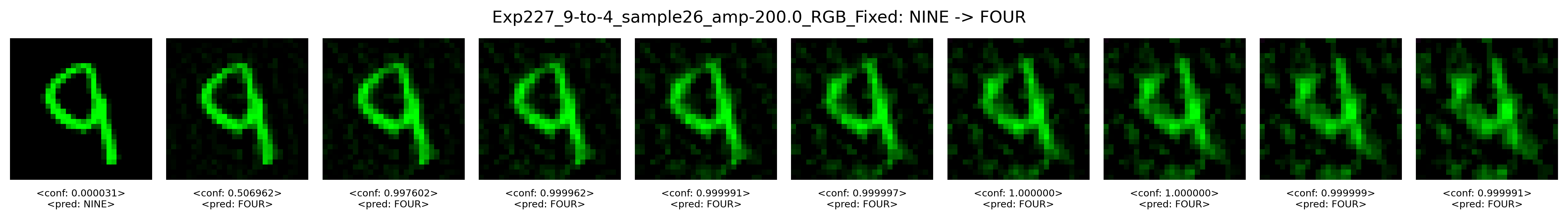}\vspace{-4mm}
    \includegraphics[width=1\textwidth, trim=0 0 0 20, clip]{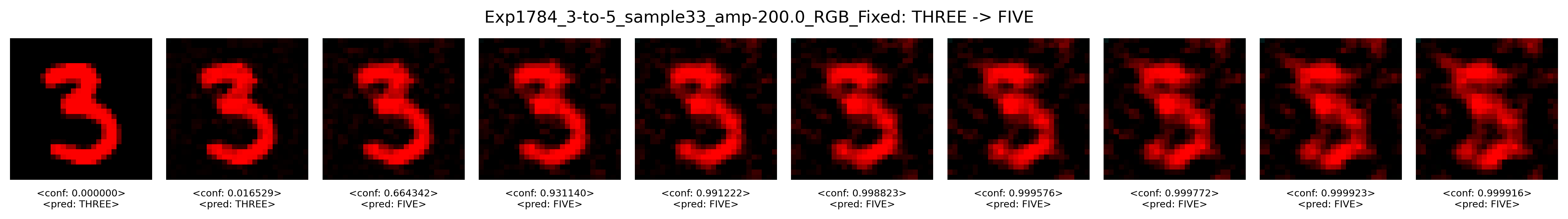}\vspace{-4mm}
    \includegraphics[width=1\textwidth, trim=0 0 0 20, clip]{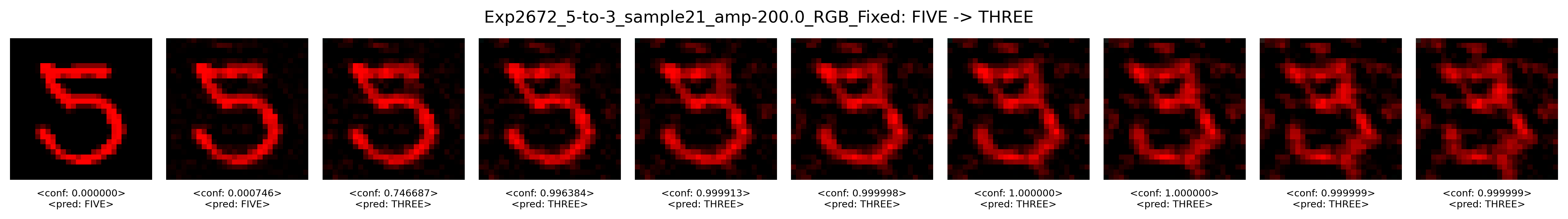}\vspace{-4mm}
    \includegraphics[width=1\textwidth, trim=0 25 0 20, clip]{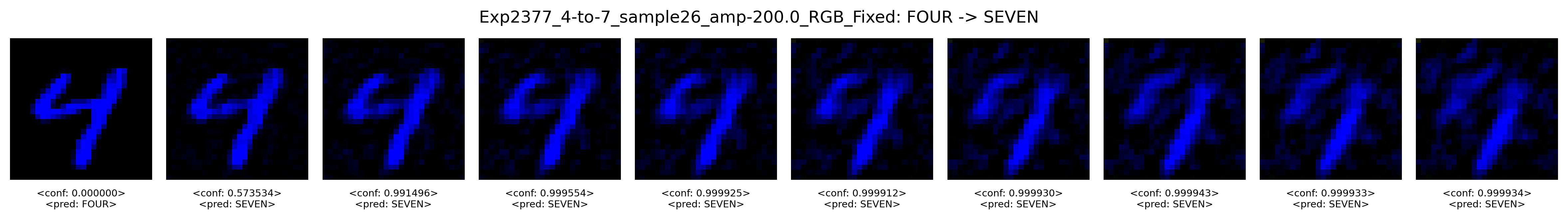}\vspace{-1mm}
     \caption{Transformation trajectories on MNIST with \emph{pure} colors in the RGB representation.  
    Each row shows a source digit morphing to the target digit of the same hue. All structural changes occur within a single RGB channel,
    which highlights the model's channel‐specific gradient steering. 
    From top to bottom, the rows transform: (5 → 3), (9 → 4), (3 → 5), (5 → 3), and (4 → 7).
    }
    
% Phase-based extrapolation on Color MNIST with pure channel colors in the RGB domain. Each row shows a source digit rendered entirely in one RGB channel (red, green, or blue) morphing to the target digit of the same hue. Because the two inactive channels remain zero, all structural changes (e.g. closing loops in ‘3’→‘8’) occur cleanly within a single color plane, highlighting the classifier’s channel-specific gradient steering.
    \label{fig:color_mnist_1}
\end{figure}

We extended our phase‐extrapolation framework to full‐color MNIST by constructing two augmented datasets. 
In the first dataset, each digit is assigned a single random color from a fixed palette, preserving the original ten classes. We observe that the classifier quickly learns to ignore color information and focus exclusively on digit shape: when asked to morph toward a target digit, the color channels remain essentially unchanged.
In the second dataset, each (digit, color) pair is treated as its own class, yielding $10 \times C$ categories. For \emph{pure} channel colors (red, green, blue), the network drives gradients exclusively in the active channel; nonactive channels stay near zero, so morphing occurs entirely within that single-color plane. For \emph{composite} hues (magenta, yellow, cyan), we extrapolate the phase independently in each of the R, G, and B channels, which allows us to steer color shifts in a controlled way. However, this requires per-channel amplifier settings, and choosing the right values can be challenging, often requiring careful, dataset-specific tuning to balance the strength of each channel update.
Figure~\ref{fig:color_mnist_1} shows example sequences. 
Direct RGB phase extrapolation often introduces visual noise. To better separate luminance from chrominance, we therefore perform extrapolation in the YIQ color space (Y = luminance; I, Q = chrominance). When morphing both digit shape and hue, the network typically adds a diffuse blob of the target color; although the classifier’s confidence in the target class increases over the trajectory, these color shifts may misalign with the digit’s contours or even desaturates its form in order to minimize loss. This behavior underscores the difficulty of jointly steering geometric structure and color information, and it motivates further exploration of perceptually coherent chrominance transforms.
 
\begin{figure}[htbp]
    \centering
    \includegraphics[width=1\textwidth, trim=0 0 0 20, clip]{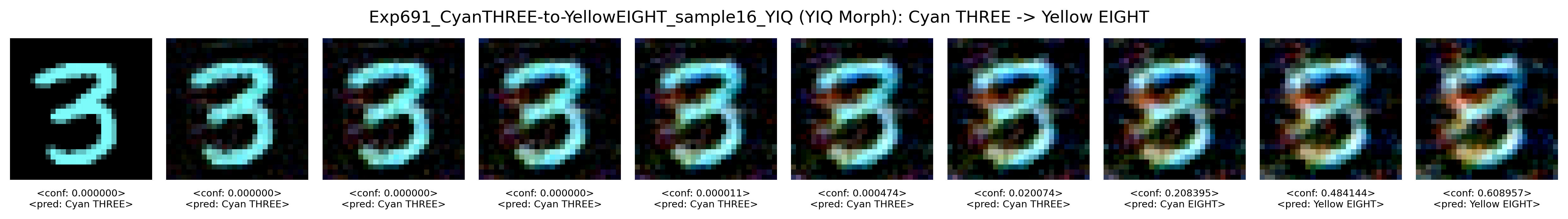}\vspace{-1mm}
    \includegraphics[width=1\textwidth, trim=0 0 0 20, clip]{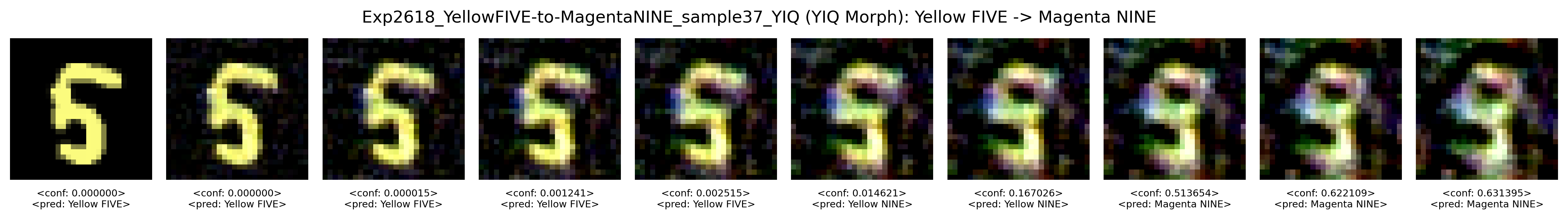}\vspace{-1mm}

    \caption{Joint color-shape morph in the YIQ representation. Target confidence increases throughout the transformation, yet color blobs may drift outside digit contours or partially desaturate the shape. The rows transform: (Cyan 3 → Yellow 8) and (Yellow 5 → Magenta 9).
    % These outcomes underscore the challenge of maintaining perceptually coherent joint control over geometry and hue.
    }
    % \caption{Joint color–shape morphs in YIQ representation. Row 1: cyan 3 → yellow 8; Row 2: yellow 5 → magenta 9. Target confidence rises even as chroma blobs sometimes stray outside the digit or partly desaturate.}
    \label{fig:color_mnist_0}
\end{figure}

\subsection{FER2013 Facial Expressions Dataset}
\begin{figure}[htbp] 
    \centering
    {\includegraphics[width=1\textwidth]{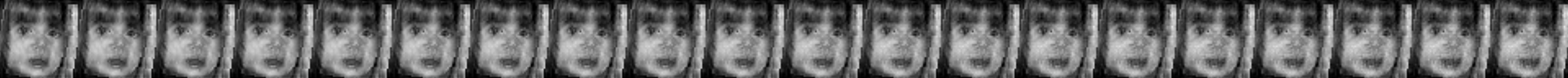}} \\
    {\includegraphics[width=1\textwidth]{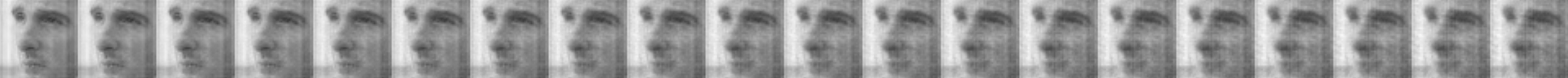}} \\
    {\includegraphics[width=1\textwidth]{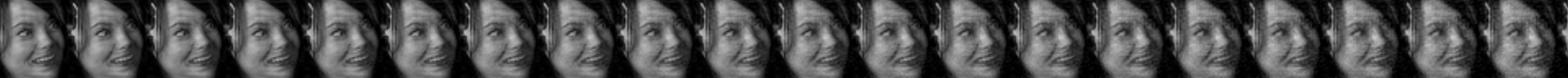}}\\
    \caption{Morphing of FER2013 samples. Images show transitions between emotions by extrapolating classifier gradients in CSP domain. Facial features like mouth shape, eye openness, and smile lines change naturally.  From top to bottom, the rows transform: (sad → happy), (sad → happy), (happy → sad).}
    \label{fig:faces}
\end{figure}

We evaluate our method on the FER2013 dataset, focusing specifically on “sad” and “happy” expressions. A CNN classifier was used to discriminate these two classes. Figure~\ref{fig:faces} shows example morphing sequences: note how the mouth curvature transitions downturned or upturned, and how smile lines emerge or soften. These extrapolations produce trajectories that align with human perception of the target emotion, highlighting the model’s sensitivity to discriminative facial cues.  

% We use the FER2013 dataset to evaluate our method on real-world emotional expressions. We focus on transformations between two distinct emotion classes of `sad' and `happy'. A CNN, incorporating channel attention and center loss for enhanced feature discriminability, was trained to classify these emotions. Figure~\ref{fig:faces} illustrates example morphing sequences generated by our method. The morphing trajectories reveal coherent and semantically meaningful changes, particularly in the mouth region, where the curvature gradually shifts between, e.g., downturned and upturned, and in the subtle emergence or softening of laugh lines. These transformations align with the perceived transition to a target emotion, demonstrating the model's sensitivities to critical class-distinguishing facial cues.
 % \caption{Visualizing inter-class transitions on FER2013. Each row depicts a morphing sequence from a source emotion (e.g., 'sad', left) to a target emotion (e.g., 'happy', right), driven by phase extrapolation of classifier gradients within the CSP representation. Note the coherent changes in key facial regions like mouth shape and eye expression, revealing the model's internal "path" between these classes.}

\section{Discussion and Conclusion}
\label{sec:conclusion}

We introduced a framework for visualizing the decision processes of neural network classifiers by generating morphing trajectories between classes. By computing gradients in transformed amplitude-phase domains and extrapolating primarily along the phase component, our method produces smooth, coherent, and interpretable sequences that illustrate how a model perceives the transition from one class to another.

\textbf{The Role of Transformed Spaces and Phase.}
The success of our approach hinges on moving away from the raw pixel representation. Pixel space is high-dimensional and highly redundant; small changes can have complex, non-local effects on semantics, and gradients are often noisy or difficult to interpret visually. Transformations like the Fourier Transform and, particularly, the Complex Steerable Pyramid (CSP)  offer representations that are more aligned with human perceptual mechanisms and the features learned by CNNs.
The CSP decomposes the image into components that are localized in space, frequency (scale), and orientation. This provides a degree of \textit{perceptual feature disentanglement} compared to pixels – a coefficient in the pyramid relates more directly to a specific visual attribute (e.g., a vertical edge at a certain location and scale) than a single pixel value does. Consequently, the relationship between these coefficients and the network's output may become locally "smoother" or more linear, making gradient extrapolation more effective and interpretable.
Our focus on \textit{phase} extrapolation is motivated by the understanding that phase in these domains primarily encodes fine-grained structural information and position, while amplitude encodes contrast or energy. Manipulating phase allows us to visualize small changes in shape, form, and arrangement, which are often crucial for class distinctions, while preserving the overall texture and contrast characteristics captured by the amplitude. This leads to morphs that appear as deformations or structural edits rather than simple brightening or blurring. The generated trajectories suggest that the network's decision manifold, when projected onto the phase dimensions of the CSP, exhibits directions corresponding to semantically meaningful transformations.

\textbf{Conclusion and Future Directions}
Our phase-based extrapolation method offers a novel lens for interpreting neural network decisions, complementing existing techniques by providing dynamic visualizations of inter-class relationships. The experiments demonstrate its ability to generate insightful morphing sequences that reveal classifier sensitivities in a structured and perceptually aligned manner.
Limitations include the reliance on the chosen transform and the assumption that a fixed gradient direction provides a meaningful path over a finite extrapolation distance. Future work could explore adaptive gradient steps, investigate alternative transforms (e.g., other wavelets, learned representations that maintain interpretability), and develop quantitative metrics to evaluate the quality and faithfulness of the generated trajectories. Extending this framework to generative models or regression tasks could also yield valuable insights. Overall, viewing classifier decisions "through a steerable lens" by analyzing gradients in structured amplitude-phase spaces appears to be a promising direction for enhancing DNN interpretability.

% \section*{Acknowledgements}
% We thank anonymous reviewers for their constructive feedback. This research was supported in part by [Funding Agency/Grant Number, if applicable].

% Bibliography
\bibliographystyle{unsrt} % Using plainnat style compatible with natbib
\bibliography{references}

%%%%%%%%%%%%%%%%%%%%%%%%%%%%%%%%%%%%%%%%%%%%%%%%%%%%%%%%%%%%
\clearpage
\appendix
\appendixpage % Creates the "Appendix" title page

\section{Wirtinger Calculus for Complex Gradients}
\label{appendix:wirtinger_derivation}

Optimizing a real-valued loss function $\mathcal{L}$ with respect to complex variables requires careful handling of gradients. Let $S \in \mathbb{C}$ be a complex variable, $S = A e^{j\Phi}$, where $A \in \mathbb{R}^+$ is the amplitude and $\Phi \in \mathbb{R}$ is the phase. Let $\mathcal{L}(S): \mathbb{C} \to \mathbb{R}$. Standard gradient descent updates the real and imaginary parts, $S = S_{re} + j S_{im}$. The update is:
\begin{align*}
S_{re, k+1} &= S_{re, k} - \alpha \frac{\partial \mathcal{L}}{\partial S_{re}} \\
S_{im, k+1} &= S_{im, k} - \alpha \frac{\partial \mathcal{L}}{\partial S_{im}}
\end{align*}
Combining these into a complex update:
\begin{align*}
S_{k+1} &= S_{re, k+1} + j S_{im, k+1} \\
&= (S_{re, k} - \alpha \frac{\partial \mathcal{L}}{\partial S_{re}}) + j (S_{im, k} - \alpha \frac{\partial \mathcal{L}}{\partial S_{im}}) \\
&= S_k - \alpha \left( \frac{\partial \mathcal{L}}{\partial S_{re}} + j \frac{\partial \mathcal{L}}{\partial S_{im}} \right)
\end{align*}
Using Wirtinger derivatives, defined as:
\[
\frac{\partial \mathcal{L}}{\partial S} = \frac{1}{2} \left( \frac{\partial \mathcal{L}}{\partial S_{re}} - j \frac{\partial \mathcal{L}}{\partial S_{im}} \right), \quad
\frac{\partial \mathcal{L}}{\partial S^*} = \frac{1}{2} \left( \frac{\partial \mathcal{L}}{\partial S_{re}} + j \frac{\partial \mathcal{L}}{\partial S_{im}} \right)
\]
where $S^*$ is the complex conjugate of $S$, we see that the gradient descent update rule is:
\begin{equation}
S_{k+1} = S_k - 2\alpha \frac{\partial \mathcal{L}}{\partial S^*}
\end{equation}
Therefore, for gradient \textit{ascent} (maximizing likelihood, minimizing negative log-likelihood) on a real loss $\mathcal{L}$ with respect to complex variables $S(\boldsymbol{\omega})$, the update direction is proportional to $\nabla_{S^*(\boldsymbol{\omega})} \mathcal{L} = \frac{\partial \mathcal{L}}{\partial S^*(\boldsymbol{\omega})}$, as used in the main text.

\section{Gradient Extrapolation Derivation in Amplitude-Phase Space}
\label{appendix:extrapolation_derivation}

We want to relate the Wirtinger gradient $\nabla_{S^*(\boldsymbol{\omega})} \mathcal{L}$ to the gradients with respect to amplitude $A(\boldsymbol{\omega})$ and phase $\Phi(\boldsymbol{\omega})$. We use the chain rule for Wirtinger derivatives. Recall $S = A e^{j\Phi}$.
\[
\frac{\partial \mathcal{L}}{\partial S^*} = \frac{\partial \mathcal{L}}{\partial A} \frac{\partial A}{\partial S^*} + \frac{\partial \mathcal{L}}{\partial \Phi} \frac{\partial \Phi}{\partial S^*}
\]
We need the derivatives of $A$ and $\Phi$ with respect to $S^*$. Since $A = \sqrt{S S^*}$ and $\Phi = \frac{1}{2j} \log(\frac{S}{S^*})$, we have:
\begin{align*}
\frac{\partial A}{\partial S^*} &= \frac{\partial}{\partial S^*} (S S^*)^{1/2} = \frac{1}{2} (S S^*)^{-1/2} S = \frac{1}{2} \frac{S}{A} = \frac{A e^{j\Phi}}{2A} = \frac{e^{j\Phi}}{2} \\
\frac{\partial \Phi}{\partial S^*} &= \frac{\partial}{\partial S^*} \left( \frac{1}{2j} (\log S - \log S^*) \right) = \frac{1}{2j} \left( -\frac{1}{S^*} \right) = \frac{j}{2S^*} = \frac{j}{2 A e^{-j\Phi}} = \frac{j e^{j\Phi}}{2A}
\end{align*}
Substituting these into the chain rule expression:
\begin{align*}
\nabla_{S^*(\boldsymbol{\omega})} \mathcal{L} &= \left( \nabla_{A} \mathcal{L} \right) \frac{e^{j\Phi}}{2} + \left( \nabla_{\Phi} \mathcal{L} \right) \frac{j e^{j\Phi}}{2A} \\
&= \frac{e^{j\Phi}}{2A} \left( A \nabla_{A} \mathcal{L} + j \nabla_{\Phi} \mathcal{L} \right) \\
&= \frac{S(\boldsymbol{\omega})}{2A(\boldsymbol{\omega})^2} \left( A(\boldsymbol{\omega}) \nabla_{A(\boldsymbol{\omega})} \mathcal{L} + j \nabla_{\Phi(\boldsymbol{\omega})} \mathcal{L} \right)
\end{align*}
Rearranging to match the form in Eq.~\eqref{eq:wirtinger_chain_rule} in the main text:
\begin{align*}
\nabla_{S^*(\boldsymbol{\omega})} \mathcal{L} &= \frac{S(\boldsymbol{\omega})}{2A(\boldsymbol{\omega})} \nabla_{A(\boldsymbol{\omega})} \mathcal{L} + \frac{j S(\boldsymbol{\omega})}{2A(\boldsymbol{\omega})^2} \nabla_{\Phi(\boldsymbol{\omega})} \mathcal{L} \\
&= \frac{1}{2} \left( \nabla_{A(\boldsymbol{\omega})} \mathcal{L} \frac{S(\boldsymbol{\omega})}{A(\boldsymbol{\omega})} + j \nabla_{\Phi(\boldsymbol{\omega})} \mathcal{L} \frac{S(\boldsymbol{\omega})}{A(\boldsymbol{\omega})^2} \right).
\end{align*}
This confirms Eq.~\eqref{eq:wirtinger_chain_rule}.

Now consider the update step $S_1 = S_0 + \alpha \nabla_{S_0^*} \mathcal{L}$. Substituting the expression for the gradient:
\begin{align*}
S_1 &= S_0 + \frac{\alpha}{2} \left( \nabla_{A_0} \mathcal{L} \frac{S_0}{A_0} + j \nabla_{\Phi_0} \mathcal{L} \frac{S_0}{A_0^2} \right) \\
&= S_0 \left( 1 + \frac{\alpha}{2} \frac{\nabla_{A_0} \mathcal{L}}{A_0} + \frac{j\alpha}{2} \frac{\nabla_{\Phi_0} \mathcal{L}}{A_0^2} \right)
\end{align*}
This matches Eq.~\eqref{eq:multiplicative_update}. Let $z = 1 + \frac{\alpha}{2} \frac{\nabla_{A_0} \mathcal{L}}{A_0} + \frac{j\alpha}{2} \frac{\nabla_{\Phi_0} \mathcal{L}}{A_0^2}$.
Then $S_1 = S_0 z = (A_0 e^{j\Phi_0}) (|z| e^{j\arg(z)}) = (|z|A_0) e^{j(\Phi_0 + \arg(z))}$.
So, $A_1 = |z| A_0$ and $\Phi_1 = \Phi_0 + \arg(z)$.
The amplitude $A_k(\boldsymbol{\omega})$ can be handled in several ways:
\begin{itemize}
    \item \textbf{Constant Amplitude:} $A_k(\boldsymbol{\omega}) = A_0(\boldsymbol{\omega})$ (adopted in this work),
    \item \textbf{Compounded Amplitude:} $A_k(\boldsymbol{\omega}) = |z|^k A_0(\boldsymbol{\omega})$,
    \item \textbf{Linear Amplitude Extrapolation:} $A_k(\boldsymbol{\omega}) = A_0(\boldsymbol{\omega})(1 + k(|z|-1))$.
\end{itemize}

The phase change for one step is $\Delta \Phi = \arg(z)$. For linear phase extrapolation over $k$ steps with constant amplitude (Option 1), we have:
\begin{align*}
\Phi_k &= \Phi_0 + k \cdot \Delta \Phi = \Phi_0 + k \cdot \arg\left( 1 + \frac{\alpha}{2} \frac{\nabla_{A_0} \mathcal{L}}{A_0} + \frac{j\alpha}{2} \frac{\nabla_{\Phi_0} \mathcal{L}}{A_0^2} \right) \\
A_k &= A_0
\end{align*}
The resulting coefficient is $S_k = A_k e^{j\Phi_k} = A_0 e^{j(\Phi_0 + k \Delta \Phi)}$.

\section{Experimental Results Using the Fourier Transform}
\label{appendix:fourier_results}
\begin{figure}[htbp] 
    \centering
    {\includegraphics[width=1\textwidth]{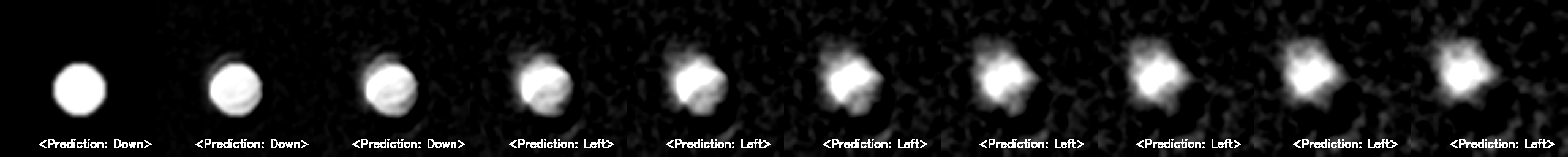}}\\
    {\includegraphics[width=1\textwidth]{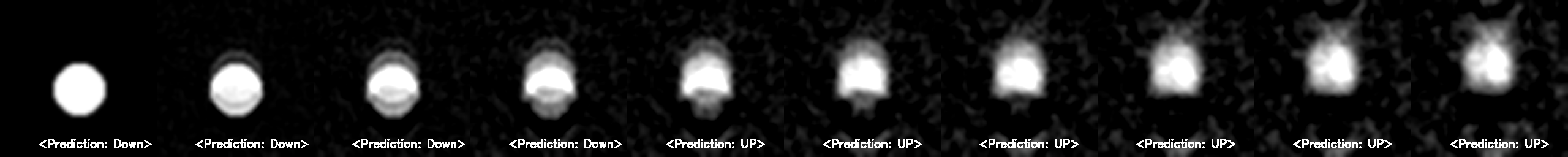}} \\
    {\includegraphics[width=1\textwidth]{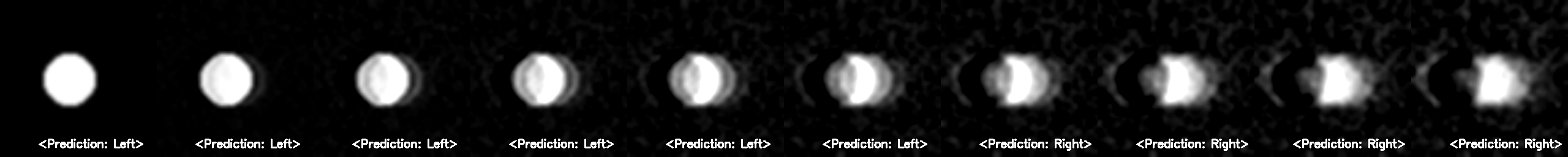}} \\
    {\includegraphics[width=1\textwidth]{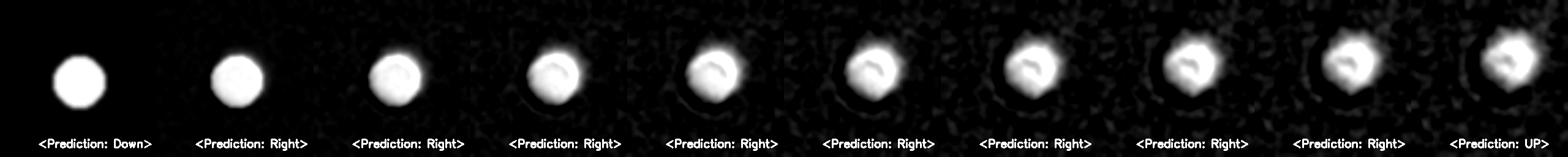}} \\
    \caption{Examples of Arcade sample transformation using Fourier Transform.}
    \label{fig:fourier_arcade_results}
\end{figure}

\begin{figure}[htbp] 
    \centering
    {\includegraphics[width=1\textwidth]{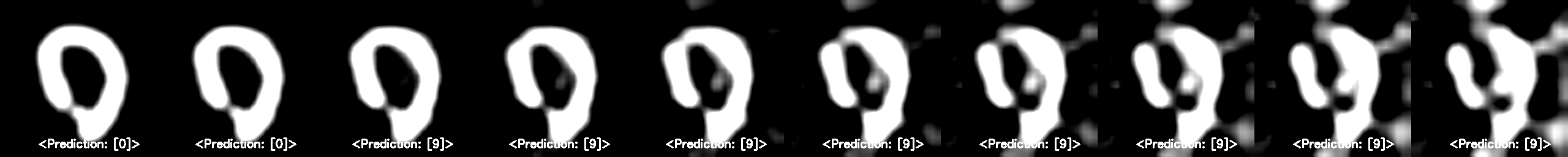}} \\
    {\includegraphics[width=1\textwidth]{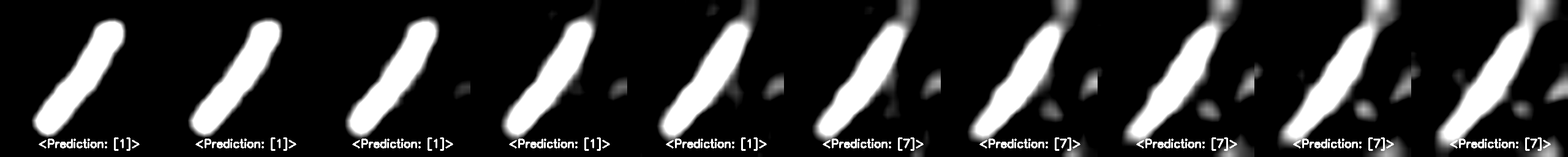}} \\
    {\includegraphics[width=1\textwidth]{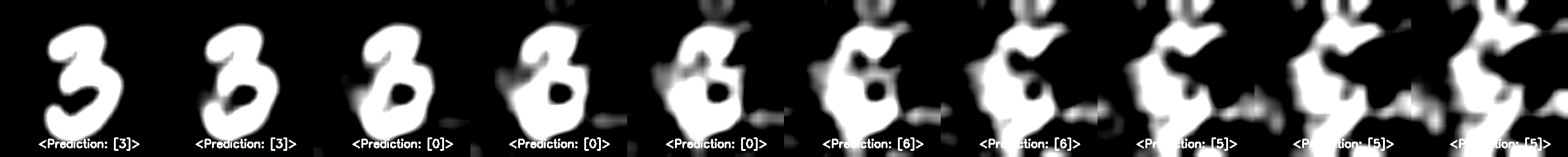}}\\
    {\includegraphics[width=1\textwidth]{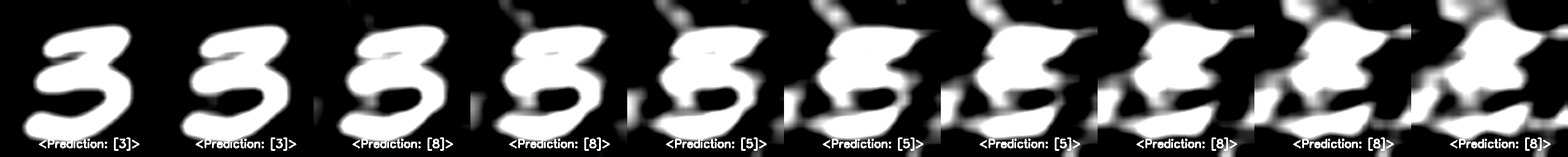}}\\
    {\includegraphics[width=1\textwidth]{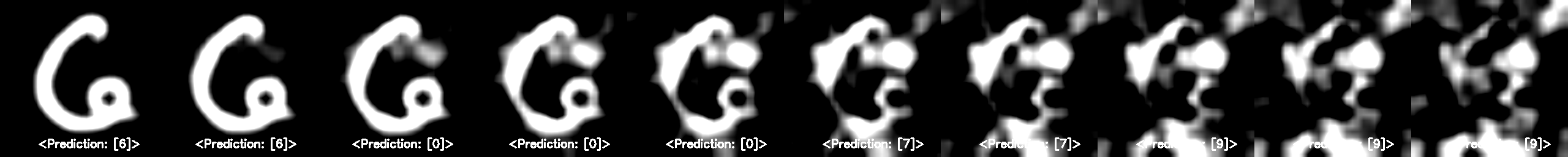}}\\
    {\includegraphics[width=1\textwidth]{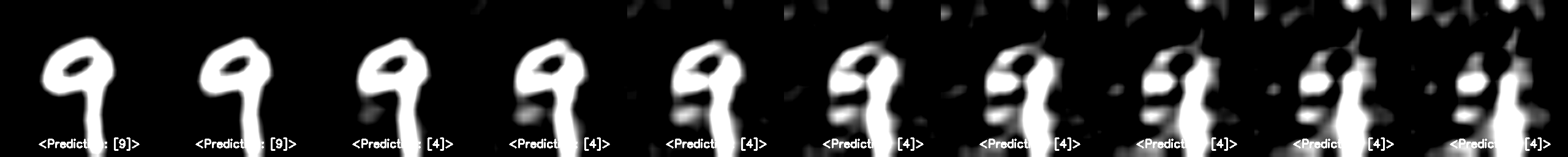}}\\
    % {\includegraphics[width=1\textwidth]{figures/329.png}}\\
    {\includegraphics[width=1\textwidth]{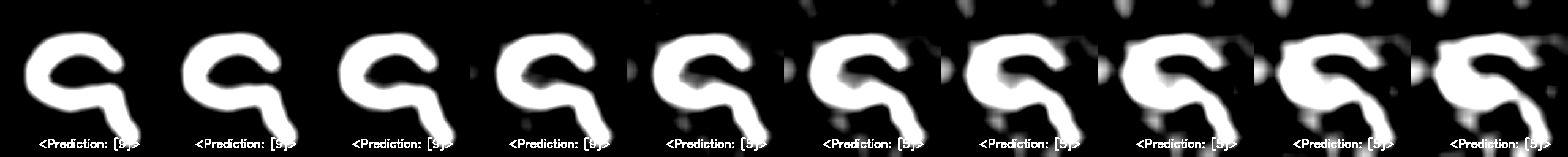}}\\
    % {\includegraphics[width=1\textwidth]{figures/429.png}}\\
    \caption{Examples of MNIST transformations using Fourier Transform phase extrapolation. Compare with Figure \ref{fig:mnist_results1}. Changes tend to be more global and less localized compared to the steerable pyramid results.}
    \label{fig:fourier_mnist_results}
\end{figure}

For completeness, we present results using the Discrete Fourier Transform (DFT) as the transformation $\mathcal{F}$. Phase extrapolation was performed using the same procedure (Eqs. \ref{eq:phase_extrap_rule}-\ref{eq:reconstruct_phase_extrap}), but operating on the DFT coefficients. Figure \ref{fig:fourier_mnist_results} shows example transformations on MNIST, while Figure \ref{fig:fourier_arcade_results} shows morphing trajectories on Arcade synthetic dataset. 

Compared to the CSP results (Figure \ref{fig:mnist_results1}), the FT-based morphs often exhibit more global changes, such as translations or large-scale warping, and can sometimes appear less sharp or artifact-prone. This is consistent with the global nature of Fourier basis functions. While still capable of transforming the digit towards the target class, the intermediate steps often lack the localized structural clarity seen with the steerable pyramid, highlighting the benefit of the CSP's localized, multi-scale, oriented representation for visualizing feature-specific sensitivities relevant to CNNs.

\end{document}